\documentclass[runningheads]{llncs}
\usepackage[misc]{ifsym}
\usepackage{graphicx}
\usepackage{amssymb}
\usepackage{amsbsy}
\usepackage{amsmath}
\usepackage{subfig}
\usepackage{comment}
\usepackage{epstopdf}
\usepackage{xcolor}
\usepackage{soul}
\usepackage[normalem]{ulem}
\usepackage{bm}
\usepackage{multirow}
\usepackage{arydshln}
\usepackage{xcolor,colortbl}
\definecolor{LightCyan}{rgb}{0.88,1,1}

\usepackage{hyperref}
\hypersetup{
    colorlinks=true,
    linkcolor=blue,
    filecolor=magenta,      
    urlcolor=magenta,
    pdftitle={Overleaf Example},
    pdfpagemode=FullScreen,
    }

\usepackage{tikz}
\usepackage{pgfplots}
\usetikzlibrary{intersections,calc,shapes,scopes,patterns,backgrounds,arrows,chains,positioning,decorations.pathreplacing,arrows.meta}

\DeclareMathOperator*{\argmin}{arg\,min}

\begin{document}

\title{Binary domain generalization for sparsifying binary neural networks\thanks{FG and MAZ are supported by the French government, through the 3IA Côte d’Azur Investments in the Future project managed by the ANR (ANR-19-P3IA-0002)}}

%
\toctitle{Binary domain generalization for sparsifying binary neural networks}
\author{Riccardo Schiavone {\Letter} \inst{1}\orcidID{0000-0002-5089-7499} \and
Francesco Galati\inst{2}\orcidID{0000-0001-6317-6298} \and
Maria~A.~Zuluaga\inst{2}\orcidID{0000-0002-1147-766X}}
\authorrunning{R. Schiavone et al.}
%
\tocauthor{Riccardo Schiavone, Francesco Galati, Maria A. Zuluaga}

\institute{Department of Electronics and Telecommunications, Politecnico di Torino, Italy\\
\email{riccardo.schiavone@polito.it} \and
Data Science Department, EURECOM, Sophia Antipolis, France \\ 
\email{\{galati,zuluaga\}@eurecom.fr} 
}
\maketitle              
\begin{abstract}
Binary neural networks (BNNs) are an attractive solution for developing and deploying deep neural network (DNN)-based applications in resource constrained devices. Despite their success, BNNs still suffer from a fixed and limited compression factor that may be explained by the fact that existing pruning methods for full-precision DNNs cannot be directly applied to BNNs. In fact, weight pruning of BNNs leads to performance degradation, which suggests that the standard binarization domain of BNNs is not well adapted for the task. This work proposes a novel more general binary domain that extends the standard  binary one that is more robust to pruning techniques, thus guaranteeing improved compression and avoiding severe performance losses. We demonstrate a closed-form solution for quantizing the weights of a full-precision network into the proposed binary domain. Finally, we show the flexibility of our method, which can be combined with other pruning strategies. Experiments over CIFAR-10 and CIFAR-100 demonstrate that the novel approach is able to generate efficient sparse networks with reduced memory usage and run-time latency, while maintaining performance.

\end{abstract}

\section{Introduction}
The increasing number of connected Internet-of-Things (IoT) devices, now surpassing the number of humans connected to the internet~\cite{evans2011}, has led to a sensors-rich world, capable of addressing real-time applications in multiple domains, where both accuracy and computational time are crucial~\cite{iot_application}. 
Deep neural networks (DNNs) have the potential of enabling a myriad of new IoT applications, thanks to their ability to process large complex heterogeneous data and to extract patterns needed to take autonomous decisions with high reliability~\cite{bengioDNN}. However, DNNs are known for being resource-greedy, 
 in terms of required computational power, memory, and energy consumption~\cite{DNNanalysis}, whereas most IoT devices are characterized by limited resources. They usually have limited processing power, small storage capabilities, they are not GPU-enabled and they are powered with batteries of limited capacity, which are expected to last over 10 years without being replaced or recharged. These constraints represent an important bottleneck towards the deployment of DNNs in IoT applications~\cite{yao2018}.

A recent and notable example to enable the usage of DNNs in limited resource devices are binary neural networks (BNNs) \cite{Courbariaux2016}. BNNs use binary weights and activation functions that allow them to replace computationally expensive multiplication operations with low-cost bitwise operations during forward propagation. This results in faster inference and better compression rates, while maintaining an acceptable accuracy for complex learning tasks~\cite{guo2022join,liu2020reactnet}. For instance, BNNs have achieved over $80\%$ classification accuracy on ImageNet~\cite{guo2022join,Imagenet}. Despite the good results, BNNs have a fixed and limited compression factor compared to full-precision DNNs, which may be insufficient for certain size and power constraints of devices~\cite{lin2020mcunet}.
 
A way to further improve BNNs' compression capacity is through network pruning, which seeks to control a network's sparsity by removing parameters and shared connections~\cite{han2015}. Pruning BNNs, however, is a more challenging task than pruning full-precision neural networks and it is still a challenge with many open questions~\cite{xu2019main}. Current attempts~\cite{guerra2021automatic,kuhar2022signed,munagala2020stq,schiavone2021sparse,wu2020sbnn,wang2021sub,xu2019main} often rely on training procedures that require more training stages than standard BNNs, making learning more complex. Moreover, these methods fail in highly pruned scenarios, showing severe accuracy degradation over simple classification problems.

In this work, we introduce sparse binary neural network (SBNN), a more robust pruning strategy to achieve sparsity and improve the performance of BNNs. Our strategy relies on entropy to optimize the network to be largely skewed to one of the two possible weight values, i.e. having a very low entropy. Unlike BNNs that use symmetric values to represent the network's weights, we propose a more general binary domain that allows the weight values to adapt to the asymmetry present in the weights distribution. This enables the network to capture valuable information, achieve better representation, and, thus better generalization. The main contributions of our work can be summarized as follows: 1) We introduce a more general binary domain w.r.t. the one used by BNNs to quantize real-valued weights; 2) we derive a closed-form solution for binary values that minimizes quantization error when real-valued weights are mapped to the proposed domain; 3) we enable the regularization of the BNNs weights distribution by using entropy constraints; 4) we present efficient implementations of the proposed algorithm, which reduce the number of bitwise operations in the network proportionally to the entropy of the weight distribution; and 5) we demonstrate SBNN's competitiveness and flexibility through benchmark evaluations.

The remaining of this work is organized as follows. Section~\ref{sec:related} discusses previous related works. The core of our contributions are described in Section~\ref{sec:method}. In Section~\ref{sec:results}, we study the properties of the proposed method and assess its performance, in terms of accuracy and operation reduction at inference, through a set of experiments using, CIFAR-10, CIFAR-100 \cite{Cifar10} and ImageNet \cite{Imagenet} datasets. Finally, a discussion on the results and main conclusions are drawn in Section~\ref{sec:conclusions}.

\section{Related Work}\label{sec:related}
We first provide an overview of BNNs. Next, we review sparsification through pruning~\cite{Bello1992,han2015,louizos2017,srivastava2014} and quantization~\cite{compress_prune,courbariaux2017,yang2019quantization,Zhang2018}, the two network compression strategies this work relies on. A broad review covering further network compression and speed-up techniques can be found in \cite{liang2021pruning}.

\vspace{0.1cm}
\noindent
\textbf{Binary Neural Networks.}
BNNs~\cite{Courbariaux2016} have gained attention in recent years due to their computational efficiency and improved compression. Subsequent works have extended \cite{Courbariaux2016} to improve its accuracy. For instance, \cite{xnornet} introduced a channel-wise scaling coefficient to decrease the quantization error. ABC-Net adopts multiple binary bases \cite{lin2017towards}, and Bi-Real \cite{liu2018birealnet} recommends short residual connection to reduce the information loss and a smoother gradient for the signum function. Recently, ReActNet~\cite{liu2020reactnet} generalized the traditional  $\text{sign}(\cdot)$ and PReLU activation functions to extend binary network capabilities, achieving an accuracy close to  full-precision ResNet-18 \cite{he2016resnet} and MobileNet V1 \cite{howard2017mobilenets} on ImageNet \cite{Imagenet}. By adopting the RSign, the RPReLU along with an attention formulation Guo et al. \cite{guo2022join} surpassed the $80\%$ accuracy mark on ImageNet. Although these works have been successful at increasing the performance of BNNs, few of them consider the compression aspect of BNNs.

\vspace{0.1cm}
\noindent
\textbf{Network Sparsification.} 
The concept of sparsity has been well studied beyond quantized neural networks as it reduces a network's computational and storage requirements and it prevents overfitting. Methods to achieve sparsity either explicitly induce it during learning through regularization (e.g. $L_{0}$ \cite{louizos2017} or $L_{1}$ \cite{han2015} regularization), or do it incrementally by gradually augmenting small networks \cite{Bello1992}; or by post hoc pruning \cite{Gomez2019,sparse_using_binary,srivastava2014}.

BNNs pruning is particularly challenging because weights in the $\lbrace\pm1\rbrace$ domain cannot be pruned based only on their magnitude. Existing methods include removing unimportant channels and filters from the network \cite{guerra2021automatic,munagala2020stq,wu2020sbnn,xu2019main}, but optimum metrics are still unclear; quantizing binary kernels to a smaller bit size than the kernel size \cite{wang2021sub}; or using the $\lbrace0,\pm 1\rbrace$ domains~\cite{kuhar2022signed,schiavone2021sparse}. Although these works suggest that the standard $\lbrace \pm1\rbrace$ binary domain has severe limitations regarding compression, BNNs using the $\lbrace0,\pm1\rbrace$ domain have reported limited generalization capabilities~\cite{kuhar2022signed,schiavone2021sparse}. In our work, we extend the traditional binary domain to a more general one, that can be efficiently implemented via sparse operations. Moreover, we address sparsity explicitly with entropy constraints, which can be formulated as magnitude pruning of the generic binary weight values mapping them in the $\lbrace0,1\rbrace$ domain. In our proposed domain, BNNs are more robust to pruning strategies and show better generalization properties than other pruning techniques for the same sparsity levels.

\vspace{0.1cm}
 \noindent
\textbf{Quantization.}
	Network quantization allows the use of fixed-point arithmetic and a smaller bit-width to represent network parameters w.r.t the full-precision counterpart. 
    Representing the values using only a finite set requires a quantization function that maps the original elements to the finite set. The quantization can be done after training the model, using parameter sharing techniques \cite{compress_prune}, or during training by quantizing the weights in the forward pass, as ternary neural networks (TNNs)~\cite{ternary2014}, BNNs~\cite{Courbariaux2015} and other quantized networks do~\cite{courbariaux2017,yang2019quantization}.
    Our work builds upon the strategy of BNNs by introducing a novel quantization function that maps weights to a binary domain that is more general than the $\{\pm1\}$ domain used in most state-of-the-art BNNs. This broader domain significantly reduces the distortion-rate curves of BNNs across various sparsity levels, enabling us to achieve greater compression.

\section{Method}\label{sec:method}
The proposed SBNN achieves network pruning via sparsification by introducing a novel quantization function that extends standard BNNs weight domain $\{\pm1\}$ to a more generic binary domain $\{\alpha,\beta\}$ and a new penalization term in the objective loss controlling the entropy of the weight distribution and the sparsity of the network (Section~\ref{problem_formulation_sec}). We derive in Section~\ref{sec:weight_optimization} the optimum SBNN's $\{\alpha,\beta\}$ values, i.e. the values that minimize the quantization loss when real-valued weights are quantized in the proposed domain. In Section~\ref{training_sec}, we use BNN's state-of-the-art training algorithms for SBNN training by adding the sparsity regularization term to the original BNN's objective loss. Section~\ref{implementation_sec} describes the implementation details of the proposed SBNN to illustrate their speed-up gains w.r.t BNNs.

\subsection{Preliminaries}\label{preliminaries}

The training of a full-precision DNN 
can be seen as a loss minimization problem: 

\begin{equation}\label{eq:standard}
\displaystyle
\argmin_{\widetilde{\textbf{W}}} \mathcal{L}(y,\hat{y})
\end{equation}
where $\mathcal{L}(\cdot)$ is a loss function between the true labels $y$ and the predicted values $\hat{y}=f(\textbf{x};\widetilde{\textbf{W}})$, which are a function of the data input $\textbf{x}$ and the network's full precision weights $\widetilde{\textbf{W}}=\smash{\lbrace \widetilde{\textbf{w}}^{\ell} \rbrace}$, with $\smash{\widetilde{\textbf{w}}^{\ell} \in \mathbb{R}^{N^{\ell}}}$ the weights of the $\ell^{th}$ layer,  and $N=\sum_{\ell} N^{\ell}$ the total number of weights in the DNN.
We denote the $i^{th}$ weight element of $\widetilde{\textbf{w}}^{\ell}$ as $\widetilde{w}_i^{\ell}$.\\

A BNN~\cite{Courbariaux2016} uses a modified signum function as quantization function that maps full precision weights $\widetilde{\textbf{W}}$ and activations $\widetilde{\textbf{a}}$ to the $\lbrace\pm 1\rbrace$ binary domain, enabling the use of low-cost bitwise operations in the forward propagation, i.e. 
\begin{equation*}
    \overline{\textbf{W}} = \text{sign}(\widetilde{\textbf{W}})\,, \qquad \dfrac{\displaystyle \partial g(\widetilde{w}_i)}{\displaystyle \partial \widetilde{w}_i} = \left\{\begin{array}{ll}
     \frac{\displaystyle \partial g(\widetilde{w}_i)}{\displaystyle \partial \overline{w}_i} & \quad \text{, if}  -1 \leq \widetilde{w}_i \leq 1 \\
     0 & \quad \text{, otherwise},
     \end{array}\right.
\end{equation*}
where $\text{sign}(\cdot)$ denotes the modified sign function over a vector, $g(\cdot)$ is a differentiable function, $\overline{\textbf{W}}$ the network's weights in the $\lbrace\pm 1\rbrace$ binary domain, $\overline{w}_i$ a given weight in the binary domain, and $\widetilde{w_i}$ the associated full-precision weight.

\subsection{Sparse Binary Neural Network (SBNN) Formulation}\label{problem_formulation_sec}
Given $\Omega^\ell=\lbrace \alpha^\ell, \beta^\ell \rbrace$ a general binary domain, with $\alpha^\ell, \beta^\ell \in \mathbb{R}$, and $\alpha^\ell < \beta^\ell$, let us define a SBNN, such that, for any given layer $\ell$,
\begin{equation}
    w_i^{\ell} \in  \Omega^\ell \qquad \forall \,\, i,
\end{equation}
with $w_i^{\ell}$ the $i^{th}$ weight element of the weight vector, ${\textbf{w}}^{\ell}$, and $\textbf{w}= \left\{\textbf{w}^{\ell}\right\}$ the set of weights for all the SBNN. 

We denote $S_{\alpha^\ell}$ and $S_{\beta^\ell}$ the indices of the weights with value $\alpha^\ell$, $\beta^\ell$ in $\mathbf{w}^\ell$
\begin{equation}
    S_{\alpha^\ell} = \lbrace i \, | \, 1 \leq i \leq N^\ell, w^\ell_i = \alpha^\ell \rbrace ,\qquad
    S_{\beta^\ell} = \lbrace i \, | \, 1 \leq i \leq N^\ell, w^\ell_i = \beta^\ell \rbrace. \nonumber
\end{equation}
Since $\alpha^\ell < \beta^\ell \,\, \forall \,\, \ell$, it is possible to estimate the number of weights taking the lower and upper values of the general binary domain over all the network:
\begin{equation}
    L^\ell = |S_{\alpha^\ell}|,\qquad U^\ell = |S_{\beta^\ell}|,\qquad L = \sum_\ell L^\ell,  \qquad U = \sum_\ell  U^\ell, \label{eq:ul2}
\end{equation}
with $L+U =N$, the total number of SBNN network weights. In the remaining of the manuscript, for simplicity and without loss of generality, please note that we drop the layer index $\ell$ from the weights notation.

To express the SBNN weights $\mathbf{w}$ in terms of binary $\lbrace 0,1 \rbrace$ weights, we now define a a mapping function  $r: \lbrace 0,1 \rbrace \longrightarrow \lbrace \alpha, \beta \rbrace$ that allows to express $\mathbf{w}$: 
\begin{equation}\label{alphabeta_to_zeroone_eq}
w_{i}  = r\left(w_{\lbrace0,1\rbrace,i}\right) = \left(w_{\lbrace0,1\rbrace,i} + \xi\right) \cdot \eta 
\end{equation}
with 
\begin{equation}
    \alpha=\xi \cdot \eta,\qquad
    \beta = (1 + \xi) \cdot \eta, \label{eq:map_alpha}
\end{equation}
 and $ w_{\lbrace0,1\rbrace,i} \in \lbrace 0,1 \rbrace$, the $i^{th}$ weight of a SBNN, when restricted to the binary set $\lbrace 0,1 \rbrace$. Through these mapping, $0$-valued weights are pruned from the network, the making SBNN sparse. 

 The bit-width of a SBNN is measured with the binary entropy $h()$ of the distribution of $\alpha$-valued and $\beta$-valued weights, 
 \begin{equation}
     h(p)= -p\log_2(p)-(1-p)\log_2(1-p) \qquad \left[ \text{bits}/ \text{weight}\right],
 \end{equation}
with $p = U / N$. Achieving network compression using a smaller bit-width than that of standard BNN's weights (1 bit/weight) is equivalent to setting a constraint in the SBNN's entropy to be less or equal than a desired value $h^*$, i.e.
\begin{equation}
    h(U/N) \leq h^*.\label{eq:entropy_constraint}
\end{equation}
Given $h^{-1}()$ the inverse binary entropy function for $0 \leq p \leq 1/2$, it is straightforward to derive such constraint,  $U\leq M$ where
\begin{equation}
    M\triangleq N \cdot h^{-1}(h^*).\label{eq:define_M}
\end{equation}
From Eq.~\eqref{eq:entropy_constraint} and \eqref{eq:define_M}, this implies that the constraint corresponds to restricting the maximum number of $1s$ in the network, and thus the sparsity of the network. Thus, the original full-precision DNN loss minimization problem (Eq.~\eqref{eq:standard}) can be reformulated as:
\begin{equation}\label{min_problem_zeroone_eq}
\begin{aligned}
\displaystyle
& \argmin_{\textbf{w}_{\lbrace0,1\rbrace},\xi,\eta}
& & \mathcal{L}(y,\hat{y}) \\
& \text{s.t.}
& & \textbf{w}_{\lbrace0,1\rbrace} \in \{0,1\}^{N}, \\
&&& U \leq M < N.
\end{aligned}
\end{equation}

The mixed optimization problem in Eq.~\eqref{min_problem_zeroone_eq} can be simplified by relaxing the sparsity constraint on $U$ through the introduction of a non-negative function $\displaystyle g(\cdot)$, which penalizes the weights when $U > M$:
\begin{equation}\label{min_problem_zeroone_loss_eq}
\begin{aligned}
\displaystyle
& \argmin_{\textbf{W}_{\lbrace0,1\rbrace},\xi,\eta}
& & \mathcal{L}(y,\hat{y}) + \lambda g(\textbf{W}_{\lbrace0,1\rbrace})\\
& \text{s.t.}
& & \textbf{W}_{\lbrace0,1\rbrace} \in \{0,1\}^{N}\\
\end{aligned}
\end{equation}
and $\lambda$ controls the influence of $g(\cdot)$. A simple, yet effective function $g(\textbf{W}_{\lbrace0,1\rbrace})$ is the following one:
\begin{equation}\label{penalty_function_eq}
g\left(\textbf{W}_{\lbrace0,1\rbrace}\right) = \text{ReLU}\left( U/N - \text{EC} \right),
\end{equation}
where $\text{EC} = M/N$ represents the fraction of expected connections, which is the fraction of $1$-valued weights in $\textbf{W}_{\lbrace0,1\rbrace}$ over the total number of weights of $\textbf{W}_{\lbrace0,1\rbrace}$.

Eq.~\eqref{min_problem_zeroone_eq} allows to compare the proposed SBNN with the standard BNN formulation. By setting $\xi= -1/2$ and $\eta = 2$, for which $\alpha = -1$ and $\beta=+1$ (Eq.~\eqref{alphabeta_to_zeroone_eq}), and removing the constraint on $U$ leads to the standard formulation of a BNN. This implies that any BNN can be represented using the $\{0,1\}$ domain and perform sparse operations. However, in practice when $U$ is not contrained to be $\leq M$, then $U\approx N/2$ and $h(1/2) = 1$ bit/weight, which means that standard BNNs cannot be compressed more.

\subsection{Weight Optimization}\label{sec:weight_optimization}
In this section, we derive the value of $\Omega=\lbrace \alpha,\beta \rbrace$ which minimizes the quantization error when real-valued weights are quantized using it.

The minimization of the quantization error accounts to minimizing the binarization loss, $\mathcal{L}_B$, which is the optimal estimator when $\widetilde{\textbf{W}}$ is mapped to $\textbf{W}$~\cite{xnornet}. This minimization is equivalent to finding the values of $\alpha$ and $\beta$ which minimize $\mathcal{L}_B$. To simplify the derivation of the optimum $\alpha$ and $\beta$ values, we minimize $\mathcal{L}_B$ over two variables in one-to-one correspondence with $\alpha$ and $\beta$. To achieve this, as in Eq.~\ref{alphabeta_to_zeroone_eq}-\ref{eq:map_alpha}, we map $w_i \in \Omega$ to $\overline{w}_i \in \{-1,+1\}$, i.e.
\begin{equation*}
   w_i = \tau\overline{w}_i+\phi, 
\end{equation*}
where $\tau$ and $\phi$ are two real-valued variables, and $\alpha=-\tau+\phi$ and $\beta=\tau+\phi$. As a result, $\alpha$ and $\beta$ are in one-to-one correspondence with $\tau$ and $\phi$, and the minimization of $\mathcal{L}_B$ can be formulated as
\begin{align}
    \begin{aligned}
    \tau^*,\phi^* & = \arg \min_{\tau,\phi} \mathcal{L}_B = \arg \min_{\tau,\phi} \left\lVert \widetilde{\textbf{w}} - (\tau\overline{\textbf{w}}+\phi\bm{1})\right\rVert_{2}
    \end{aligned} \label{eq:minimization_phitau}
\end{align}
where $\lVert\cdot\rVert_{2}$ is the $\ell_2$-norm and $\bm{1}$ is the all-one entries matrix.

By first expanding the $\ell_2$-norm term and using the fact that $\text{sum}(\overline{\textbf{w}}) = N^{\ell} (2p-1)$, it is straightforward to reformulate Eq.~\ref{eq:minimization_phitau} as a a function of the sum of real-valued weights, their ${\ell_1}$-norm, the fraction of $+1$-valued binarized weights and the two optimization parameters. In such case, the $\nabla \mathcal{L}_B$ is 
\begin{equation}\label{eq:gradient}
    \nabla \mathcal{L}_B = \begin{pmatrix}
      \frac{\partial \mathcal{L}_B}{\partial \tau}\\
      \frac{\partial \mathcal{L}_B}{\partial \phi}
    \end{pmatrix} = 2 \begin{pmatrix}-\lVert\widetilde{\textbf{w}}\rVert_{1}+N^{\ell}\bigl(\tau  + \phi (2p-1)\bigr)\\
    -\text{ sum}(\widetilde{\textbf{w}})+N^{\ell} \bigl(\phi + \tau (2p-1)\bigr)
    \end{pmatrix}.
\end{equation}
Solving to find the optimal values $\tau$ and $\phi$ we obtain
\begin{equation}
    \tau^* = \frac{\lVert\widetilde{\textbf{w}}\rVert_{1}}{N^{\ell}} - \phi^* (2p-1)\,,\,\,\,\,\, \phi^* = \frac{\text{sum}(\widetilde{\textbf{w}})}{N^{\ell}} - \tau^* (2p-1). \label{eq:optimal_phitau}
\end{equation}
When $p=0.5$, like in standard BNNs, it gives the classical value of $\tau^*=\lVert\widetilde{\textbf{w}}\rVert_{1}/N^{\ell}$ as in \cite{xnornet}.
By substituting $\phi^*$ in Eq. \eqref{eq:minimization_phitau}, we obtain the closed-form solution 
\begin{equation}
    \tau^* = \frac{\lVert\widetilde{\textbf{w}}\rVert_{1}-(2p-1)\text{sum}(\widetilde{\textbf{w}})}{N^{\ell}(1- (2p-1)^2)}\,,\,\,\,\,\,
    \phi^* = \frac{\text{ sum}(\widetilde{\textbf{w}})-(2p-1)\lVert\widetilde{\textbf{w}}\rVert_{1}}{N^{\ell}(1- (2p-1)^2)}. \label{eq:optimal_phitau_closed}
\end{equation}

As the gradient (Eq.~\ref{eq:gradient}) is linear in $\phi$ and $\tau$, this implies that there is a unique critical point. Moreover, an analysis of the Hessian matrix confirms that $\mathcal{L}_B$ is convex and that local minimum is a global minimum. The derivation is here omitted as it is straightforward. 

\subsection{Network Training}\label{training_sec}
The SBNN training algorithm builds upon state-of-the-art BNN training algorithms~\cite{bethge2019backtosimplicity,Courbariaux2016,liu2020reactnet}, while introducing network sparsification. To profit from BNNs training scheme, we replace $\mathbf{W}_{\lbrace0,1\rbrace}, \xi$ and $\eta$ (Eq. \eqref{min_problem_zeroone_loss_eq}) with $\overline{W}, \tau$ and $\phi$. Doing so, $\mathcal{L}(y,\hat{y})$ corresponds to the loss of BNN algorithms $\mathcal{L}_{\text{\tiny{BNN}}}$. SBNN training also requires to add the penalization term from Eq.~\eqref{penalty_function_eq} to account for sparsity. To account for $\overline{\textbf{W}}$, the regularization function $g(\mathbf{W}_{\lbrace0,1\rbrace})$ (Eq.~\eqref{penalty_function_eq}) is redefined according to
\begingroup\makeatletter\def\f@size{9.5}\check@mathfonts
\begin{equation}\label{actual_ones_eq}
j(\overline{\textbf{W}}) = \text{ReLU}\left(\left(\sum_i \dfrac{\overline{w}_i +1}{2N}\right)-\text{EC}\right), 
\end{equation}
\endgroup
and the SBNN objective loss can be expressed as 
\begin{equation}\label{loss_bnn_sbnn_eq}
  \mathcal{L}_{\text{\tiny{SBNN}}} = \mathcal{L}_{\text{\tiny{BNN}}} + \lambda \,j(\overline{\textbf{W}}).
\end{equation}

During training, we modulate the contribution of the regularization term $j(\overline{\textbf{W}})$ by imposing, at every training iteration, to be equal to a fraction of $\mathcal{L}_{\text{\tiny{SBNN}}}$, i.e.
\begin{equation}\label{gamma_eq}
  \gamma = \dfrac{\lambda\,j(\overline{\textbf{W}})}{\mathcal{L}_{\text{\tiny{SBNN}}}}.
\end{equation}
The hyperparameter $\gamma$ is set to a fixed value over all the training process. Since $\mathcal{L}_{\text{\tiny{SBNN}}}$ changes at every iteration, this forces $\lambda$ to adapt, thus modulating the influence of $j(\overline{\textbf{W}})$ proportionally to the changes in the loss. The lower $\gamma$ is set, the less influence $j(\overline{\textbf{W}})$ has on the total loss. This means that network sparsification will be slower, but convergence will be achieved faster. On the opposite case (high $\gamma$), the training will favor sparsification.

\begin{figure*}[t]
 \centering		\includegraphics[clip,trim={0cm, 0cm, 8cm, 0cm},width=\textwidth]{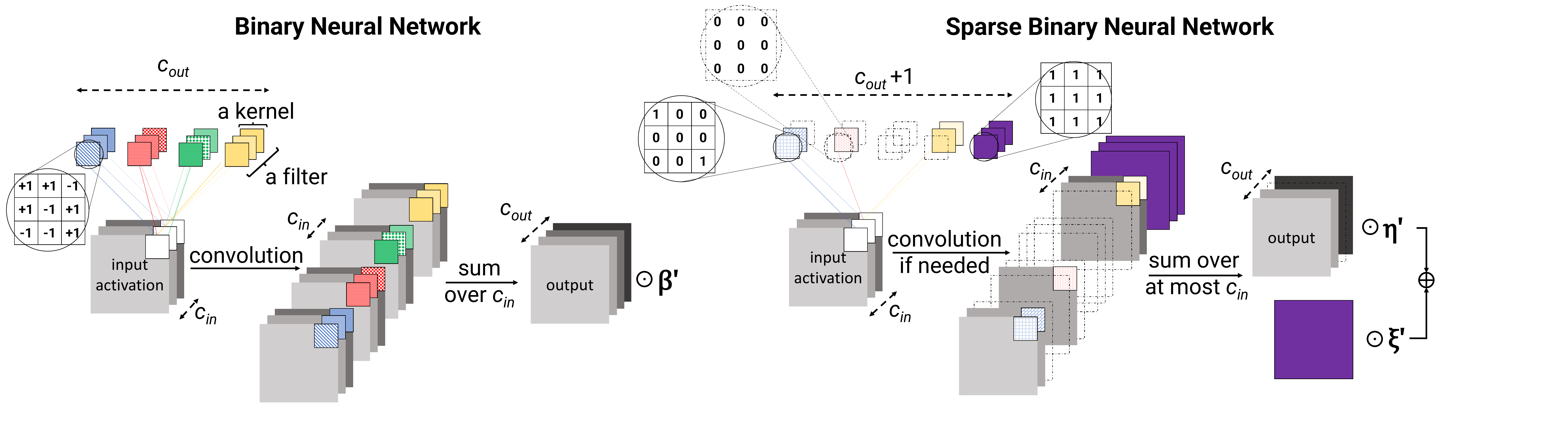}
					
		\caption{BNNs vs. SBNNs operations in a convolutional layer  using $c_{out}$ filters and input of $c_{in}$ dimensions. BNNs' $(c_{out} \cdot c_{in})$ convolutional kernels are dense and require all computations. SBNNs' kernels are sparse, allowing to skip certain convolutions and sum operations. The removed filters are indicated by a dashed contour and no fill. Both BNNs and SBNNs perform convolutions using XNOR and popcount operations, while the sum is replaced by popcount operations.}
		\label{implementation_fig}
	\end{figure*}

 \subsection{Implementation Gains}\label{implementation_sec}
We discuss the speed-up gains of the proposed SBNN through its efficient implementation using linear layers in the backbone architecture. Its extension to convolutional layers (Fig.~\ref{implementation_fig}) is straightforward, thus we omit it for the sake of brevity. 

We describe the use of sparse operations, as it can be done on an FPGA device~\cite{fu2022towards,wang2021sub}. Instead, when implemented on CPUs, SBNNs can take advantage of pruned layers, kernels and filters for acceleration~ \cite{guerra2021automatic,munagala2020stq,wu2020sbnn,xu2019main}. Moreover, for kernels with only a single binary weight equal to $1$ there is no need to perform a convolution, since the kernels remove some elements from the corner of their input.

The connections in a SBNN are the mapped one-valued weights, i.e. the set $S_1$. Therefore, SBNNs do not require any $\mathrm{XNOR}$ operation on FPGA, being $\mathrm{popcount}$ the only bitwise operation needed during the forward pass. The latter, however, is performed only in a layer's input bits connected through the one-valued weights rather than the full input.

For any given layer $\ell$, the number of binary operations of a BNN is $\mathcal{O}_{\text{\tiny{BNN}}} = 2 N^{\ell}$ \cite{bethge2019backtosimplicity}, $N^{\ell}$ $\mathrm{XNOR}$ operations and $N^{\ell}$ $\mathrm{popcount}$s. A rough estimate of the implementation gain in terms of the number of binary operations of SBNNs w.r.t. BNNs can be expressed in terms of the EC as
\begin{equation}
  \dfrac{\mathcal{O}_{\text{\tiny{SBNN}}}}{\mathcal{O}_{\text{\tiny{BNN}}}} \approx \dfrac{2N^{\ell}}{\text{EC}\cdot N^{\ell}} \approx \dfrac{2}{\text{EC}},
\end{equation}
which indicates that the lower the EC fraction, the higher the gain w.r.t. BNNs.

Binary operations are not the only ones involved in the inference of SBNN layers. After the sparse $\{0,1\}$ computations, the mapping operations to the $\{\alpha,\beta\}$ domain take place, also benefiting from implementation gains. To analyze these, let us now denote $\mathbf{x}$ the input vector to any layer and $\mathbf{z}=\mathbf{w}\,\mathbf{x}$ its output. Using E. \eqref{alphabeta_to_zeroone_eq}, $\mathbf{z}$ can be computed as
\begin{equation} 
\mathbf{z} = \xi\,\mathbf{z}' + \xi\,\eta\,\mathbf{q},
\label{implementation_eq}
\end{equation}
where $\mathbf{z}'=\mathbf{w_{\lbrace0,1\rbrace}}\,\mathbf{x}$ is the result of sparse operations (Fig.~\ref{implementation_fig}), $\mathbf{q} = \mathbf{1}\,\mathbf{x}$, and $\mathbf{1}$ the all-ones matrix.

All the elements in $\mathbf{q}$ take the value $2\cdot\mathrm{popcount}(\mathbf{x}) - |\mathbf{x}|$, with $|\mathbf{x}|$ the size of $\mathbf{x}$. Therefore, they are computed only once, for each row of $\mathbf{1}$. Being $\xi$ and $\eta$ known at inference time, they can be used to precompute the threshold in the threshold comparison stage of the implementation of the $\mathrm{batchnorm}$ and $\mathrm{sign}$ operations following the estimation of $\mathbf{z}$~\cite{umuroglu2017finn}. Thus, SBNNs require $|\mathbf{x}|$ binary operations, one real product and $|\mathbf{x}|$ real sums to obtain $\mathbf{z}$ from $\mathbf{z}'$.

\section{Experiments and Results}\label{sec:results}
We first run a set of ablation studies to analyze the properties of the proposed method (Section \ref{sec:result_domain_comparison}). Namely, we analyze the generalization of SBNNs in a standard binary domain and the proposed generic binary domain; we study the role of the quantization error in the network's performance; and the effects of sparsifying binary kernels. 
Next, we compare our proposed method to other state-of-the-art techniques using the well established CIFAR-10 and CIFAR-100~\cite{Cifar10} datasets. Preliminary results on ImageNet~\cite{Imagenet} are also discussed. 
All our code has been made publicly available\footnote{\href{https://github.com/robustml-eurecom/SBNN/}{github.com/robustml-eurecom/SBNN}}.

\subsection{Ablation Studies}\label{sec:result_domain_comparison}
\noindent
\textbf{Experimental setup.} We use a ResNet-18 binarized model trained on CIFAR-10 as backbone architecture. We train the networks for 300 epochs, with batch size of 512, learning rate of $1e-3$, and standard data augmentation techniques (random crops, rotations, horizontal flips and normalization). We use an Adam optimizer and the cosine annealer for updating the learning rate as suggested in \cite{liu2021adam} and we follow the binarization strategy of IR-Net \cite{qin2020forward}.\\

\noindent
\textbf{Generalization properties.}\label{sec:comparison_domains} We compare the performance of the proposed generic binary domain to other binary domains used by BNNs by assessing the networks' generalization capabilities when the sparsity ratio is $95\%$. For this experiment, 
we use the $\{-\beta,+\beta\}$ domain from~\cite{xnornet} with no sparsity constraints as the baseline. Additionally, we consider the same domain with a $95\%$ sparsity constraint and the $\{\alpha,\beta\}$ domain obtained optimizing $\tau$ and $\phi$ according to Eq. \eqref{eq:optimal_phitau_closed} with the $95\%$ sparsity constraint.
Table \ref{tab:quantization_error} reports the obtained results in terms of top-1 accuracy and accuracy loss w.r.t. the BNN baseline model ($\Delta$). When we impose the $95\%$ sparsity constraint with the $\{-\beta,+\beta\}$ domain, the accuracy drop w.r.t. to the baseline is $2.98\%$. Using the $\{\alpha,\beta\}$ domain,
 the loss goes down to $2.47\%$, nearly $0.5\%$ better than the $\{-\beta,+\beta\}$ domain. The results suggest that a more general domain leads to improved generalization capabilities.\\ 

\noindent
\textbf{Impact of the quantization error}\label{sec:quantization_error}
We investigate the impact of the quantization error in the SBNN generalization. To this end, we compare the proposed quantization technique (Sec.~\ref{sec:weight_optimization}) with the strategy of learning $\Omega$ via back-propagation.
We denote this approach Learned $\{\alpha,\beta\}$ (Table~\ref{tab:quantization_error}).
The obtained results show that with the learning of the parameters the accuracy loss w.r.t. the BNN baseline decreases down to $-0.09\%$, thus $2.38\%$ better than when $\tau$ and $\phi$ are analytically obtained with Eq.~\eqref{eq:optimal_phitau_closed}. This result implies that the quantization error is one of the sources of accuracy degradation when mapping real-valued weights to any binary domain, but it is not the only source. Indeed, activations are also quantized. Moreover, errors are propagated throughout the network. Learning $\Omega$ can partially compensate for these other error sources.

\begin{table*}[t]
	\caption{Role of the binary domain and the quantization error when sparsifying BNNs. Experiments performed on CIFAR-10 with a binarized ResNet-18 model.} 
		\label{tab:quantization_error}
		\begin{center}
    {%
			\begin{tabular}{c|ccc}
			     \hline 
			     Domain & Sparsity constraint & Top-1 Accuracy & $\Delta$ \\\hline \hline
                    Baseline & / & 88.93\% & /\\\cdashline{1-4}
                     $\{-\beta,+\beta\}$ \cite{xnornet} & 95\% & 85.95\% & -2.98\%\\
                     $\{\alpha,\beta\}$ & 95\% & 86.46\% & -2.47\%\\
                     Learned $\{\alpha,\beta\}$ & 95\% & 88.84\% & -0.09\%\\
                     \hline
                \end{tabular}
			}
		\end{center}
  \vspace{-0.75cm}
	\end{table*}

 \ \\
\noindent
 \textbf{Effects of network sparsification}\label{sec:pruning_effects}
We investigate the effects of network sparsification and how they can be leveraged to reduce the binary operations (BOPs) required in SBNNs. In Section \ref{sec:comparison_domains}, we showed that our binary domain is more adept at learning sparse network representations compared to the standard binary domain. This allows us to increase the sparsity of SBNNs while maintaining a desired level of accuracy. When the sparsity is sufficiently high, many convolutional kernels can be entirely removed from the network, which further reduces the BOPs required for SBNNs. Additionally, convolutional kernels with only a single binary weight equal to $1$ do not require a convolution to be performed, as these kernels simply remove certain elements from the input.

To illustrate this effect, we plotted the distribution of binary kernels for the $5$th, $10$th, and $15$th layers of a binarized ResNet-18 model (Fig. ~\ref{tab:kernels}). The first column shows the distribution when no sparsity constraints are imposed, while the second and third columns show the distribution for sparsity levels of $95\%$ and $99\%$, respectively. The kernels are grouped based on their Hamming weights, which is the number of non-zero elements in each $\{0,1\}^{3\times3}$ kernel. The plots suggest that increasing the sparsity of SBNNs results in a higher number of kernels with Hamming weights of $0$ and $1$.

\newcommand\w{0.39}
\newcommand\h{0.22}

    \begin{table}[!t]
        \centering
        \begin{tabular}{cccc}
            & Sparsity $50\%$ & Sparsity $95\%$ & Sparsity $99\%$ \\
            \multirow{3}{*}{\rotatebox[origin=c]{90}{Percentage of kernels}} & \begin{tikzpicture}
\begin{axis}[name=plot1, xmin = -0.25, xmax = 9.25, ymin = 0, ymax = 15, xtick= {0,1,2,3,4,5,6,7,8,9}, ytick={0,5,10,15,20,30,40,50,60,70,80,90,100}, 
grid = both, minor tick num = 0, major grid style = {lightgray},   minor grid style = {lightgray!25}, width = \w\columnwidth, height = \h\columnwidth]
    \addplot+ [ycomb,] file {Graphics/results_kernels/result_EC50_L5_kernelsweights.dat};
\end{axis}
\end{tikzpicture} & \begin{tikzpicture}
\begin{axis}[name=plot2, at=(plot1.right of south east), xmin = -0.25, xmax = 9.25, ymin = 0, ymax = 60, xtick= {0,1,2,3,4,5,6,7,8,9}, ytick={0,20,40,60,70,80,90,100}, 
grid = both, minor tick num = 0, major grid style = {lightgray},   minor grid style = {lightgray!25}, width = \w\columnwidth, height = \h\columnwidth]
    \addplot+ [ycomb,] file {Graphics/results_kernels/result_EC5_L5_kernelsweights.dat};
\end{axis}
\end{tikzpicture} & \begin{tikzpicture}
\begin{axis}[name=plot3, at=(plot2.right of south east), xmin = -0.25, xmax = 9.25, ymin = 0, ymax = 100, xtick= {0,1,2,3,4,5,6,7,8,9}, ytick={0,30,60,90}, 
grid = both, minor tick num = 0, major grid style = {lightgray},   minor grid style = {lightgray!25}, width = \w\columnwidth, height = \h\columnwidth]
    \addplot+ [ycomb,] file {Graphics/results_kernels/result_EC1_L5_kernelsweights.dat};
\end{axis}
\end{tikzpicture} \\
             & \begin{tikzpicture}
\begin{axis}[name=plot1, xmin = -0.25, xmax = 9.25, ymin = 0, ymax = 20, xtick= {0,1,2,3,4,5,6,7,8,9}, ytick={0,5,10,15,20,30,40,50,60,70,80,90,100}, 
grid = both, minor tick num = 0, major grid style = {lightgray},   minor grid style = {lightgray!25}, width = \w\columnwidth, height = \h\columnwidth]
    \addplot+ [ycomb,] file {Graphics/results_kernels/result_EC50_L10_kernelsweights.dat};
\end{axis}
\end{tikzpicture} & \begin{tikzpicture}
\begin{axis}[name=plot2, at=(plot1.right of south east), xmin = -0.25, xmax = 9.25, ymin = 0, ymax = 60, xtick= {0,1,2,3,4,5,6,7,8,9}, ytick={0,20,40,60,70,80,90,100}, 
grid = both, minor tick num = 0, major grid style = {lightgray},   minor grid style = {lightgray!25}, width = \w\columnwidth, height = \h\columnwidth]
    \addplot+ [ycomb,] file {Graphics/results_kernels/result_EC5_L10_kernelsweights.dat};
\end{axis}
\end{tikzpicture} & \begin{tikzpicture}
\begin{axis}[name=plot3, at=(plot2.right of south east), xmin = -0.25, xmax = 9.25, ymin = 0, ymax = 100, xtick= {0,1,2,3,4,5,6,7,8,9}, ytick={0,30,60,90}, 
grid = both, minor tick num = 0, major grid style = {lightgray},   minor grid style = {lightgray!25}, width = \w\columnwidth, height = \h\columnwidth]
    \addplot+ [ycomb,] file {Graphics/results_kernels/result_EC1_L10_kernelsweights.dat};
\end{axis}
\end{tikzpicture} \\
           & \begin{tikzpicture}
\begin{axis}[name=plot1, xmin = -0.25, xmax = 9.25, ymin = 0, ymax = 15, xtick= {0,1,2,3,4,5,6,7,8,9}, ytick={0,5,10,15,20,30,40,50,60,70,80,90,100}, 
grid = both, minor tick num = 0, major grid style = {lightgray},   minor grid style = {lightgray!25}, width = \w\columnwidth, height = \h\columnwidth]
    \addplot+ [ycomb,] file {Graphics/results_kernels/result_EC50_L15_kernelsweights.dat};
\end{axis}
\end{tikzpicture} & \begin{tikzpicture}
\begin{axis}[name=plot2, at=(plot1.right of south east), xmin = -0.25, xmax = 9.25, ymin = 0, ymax = 100, xtick= {0,1,2,3,4,5,6,7,8,9}, ytick={0,30,60,90}, 
grid = both, minor tick num = 0, major grid style = {lightgray},   minor grid style = {lightgray!25}, width = \w\columnwidth, height = \h\columnwidth]
    \addplot+ [ycomb,] file {Graphics/results_kernels/result_EC5_L15_kernelsweights.dat};
\end{axis}
\end{tikzpicture} & \begin{tikzpicture}
\begin{axis}[name=plot3, at=(plot2.right of south east), xmin = -0.25, xmax = 9.25, ymin = 0, ymax = 105, xtick= {0,1,2,3,4,5,6,7,8,9}, ytick={0,33,67,100}, 
grid = both, minor tick num = 0, major grid style = {lightgray},   minor grid style = {lightgray!25}, width = \w\columnwidth, height = \h\columnwidth]
    \addplot+ [ycomb,] file {Graphics/results_kernels/result_EC1_L15_kernelsweights.dat};
\end{axis}
\end{tikzpicture} \\
& & kernel's Hamming weight & \\
        \end{tabular}
        
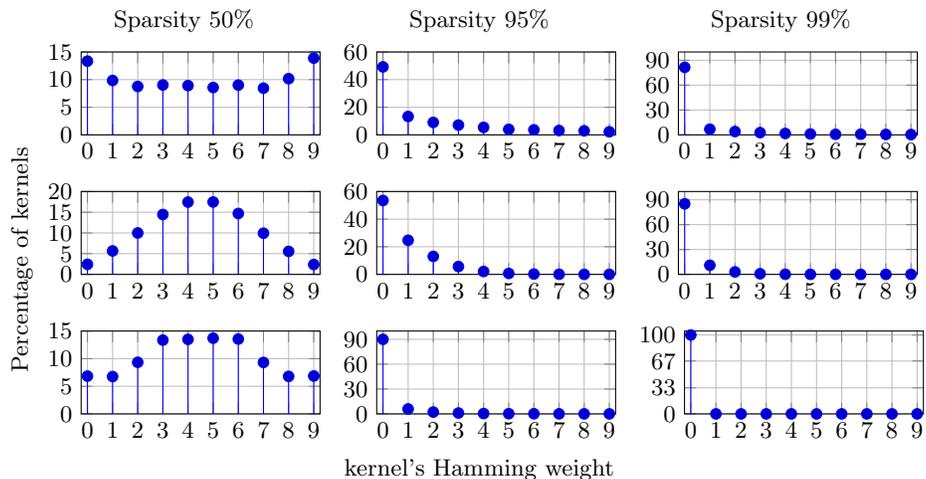
\captionof{figure}{Percentage of binary kernels for  various Hamming weights of a binarized Resnet-18 model over CIFAR-10 for different sparsity constraints. The $5$-th, $10$-th and $15$-th layers are shown in the top, middle and bottom rows, respectively.}
        \label{tab:kernels}
        \vspace{-0.75cm}
    \end{table}

\subsection{Benchmark}\label{sec:benchmark}
\noindent
\textbf{CIFAR-10.}
We compare our method against state-of-the-art methods over a binarized ResNet-18 model using CIFAR-10. Namely, we consider: STQ \cite{munagala2020stq}, Slimming \cite{wu2020sbnn}, Dual-P \cite{fu2022towards}, Subbit \cite{wang2021sub}, IR-Net~\cite{qin2020forward} and our method with learned $\tau$ and $\phi$, for different sparsity constraints. We use the IR-Net as BNN baseline to be compressed.
We use the experimental setup described in Sec. \ref{sec:result_domain_comparison} with some modifications. We extend the epochs to 500 as in~\cite{wang2021sub}, and we use a MixUp strategy~\cite{zhang2017mixup}. In the original IR-Net formulation~\cite{qin2020forward}, the training setup is missing. We use our setup to train it, achieving the  same accuracy as in \cite{qin2020forward}.

Table \ref{tab:CIFAR_10} reports the obtained results in terms of accuracy (Acc.), accuracy loss w.r.t. the IR-Net model ($\Delta$), and BOPs reduction (BOPs PR). For our SBNN, we estimate BOPs PR by 
counting the number of operations which are not computed from the convolutional kernels with Hamming weight $0$ and $1$. For other methods, we refer the reader to the original publications. 
We assess our method at different levels of sparsity, in the range 50 to 99\%. For SBNNs we also report  the percentage of SBNN's convolutional kernels with Hamming weight $0$ ($K_0$) and with Hamming weight $1$ ($K_1$).

The results suggest that our method is competitive with other more complex pruning strategies. Moreover, our method reports similar accuracy drops w.r.t. state-of-the-art Subbit and Dual-P for similar BOPs PR. However, we need to point out that Subbit and Dual-P results refer to BOPs PR on FPGA, where SBNN can take advantage of sparse operations (Section \ref{implementation_sec}) also for the kernels with larger Hamming weights than $0$ and $1$, because on FPGA all operations involving $0$-valued weights can be skipped. For instance, the use of sparse operations on the SBNN 95\% allows to remove $\approx$ 84.9\% BOPs.

\begin{table}[t]
\scriptsize
		\caption{Evaluation of kernel removal for different pruning targets using a binarized Resnet-18 model on CIFAR-10.} 
		\label{tab:CIFAR_10}
		\begin{center}
		  \resizebox{.8\linewidth}{!}{%
			\begin{tabular}{ccccrr}
			     \hline
			     \multicolumn{1}{c}{\textbf{Method}} & \multicolumn{1}{c}{\textbf{Acc.}} & \multicolumn{1}{c}{$\bm{\Delta}$} & \multicolumn{1}{c}{\textbf{BOPs PR}} &\multicolumn{1}{c}{$\bm{K_0}$} & \multicolumn{1}{c}{$\bm{K_1}$}  \\\hline
                    IR-Net  & 91.50\% & / & / & / & /\\\cdashline{1-6}
                    STQ  & 86.56\% & -5.50\% & -40.0\%& / & / \\
                    Slimming  & 89.30\% & -2.20\% & -50.0\%& / & / \\
                    Dual-P (2$\rightarrow$1)  & 91.02\% & -0.48\% & -70.0\%& / & /\\
                    Dual-P (3$\rightarrow$1)  & 89.81\% & -1.69\% & -80.6\%& / & / \\
                    Dual-P (4$\rightarrow$1)  & 89.43\% & -2.07\% & -85.4\% & / & / \\
                    Subbit $0.67$-bits   & 91.00\% & -0.50\% & -47.2\%& / & / \\
                    Subbit $0.56$-bits  & 90.60\% & -0.90\%  & -70.0\% & / & /\\
                    Subbit $0.44$-bits & 90.10\% & -1.40\% & -82.3\%& / & /\\
                     \rowcolor{LightCyan}
                     SBNN 50\% [\textbf{our}]  & 91.70\% & +0.20\% & -11.1\%& 5.6\% & 6.8\% \\
                     \rowcolor{LightCyan}
                     SBNN 75\% [\textbf{our}]  & 91.71\% & +0.21\% & -24.5\% & 30.7\% & 15.9\% \\
                     \rowcolor{LightCyan}
                     SBNN 90\% [\textbf{our}]  & 91.16\% & -0.24\% & -46.5\%& 61.8\% & 15.5\% \\
                     \rowcolor{LightCyan}
                     SBNN 95\% [\textbf{our}]  & 90.94\% & -0.56\% & -63.2\%& 77.1\% & 11.8\% \\
                     \rowcolor{LightCyan}
                     SBNN 96\% [\textbf{our}]  & 90.59\% & -0.91\% & -69.7\%& 81.0\% & 10.1\% \\
                     \rowcolor{LightCyan}
                     SBNN 97\% [\textbf{our}]  & 90.71\% & -0.79\%  & -75.7\%& 84.8\% & 8.7\%\\
                     \rowcolor{LightCyan}
                     SBNN 98\% [\textbf{our}]  & 89.68\% & -1.82\%  & -82.5\%& 89.3\% & 6.5\%\\
                     \rowcolor{LightCyan}
                     SBNN 99\% [\textbf{our}]  & 88.87\% & -2.63\%  & -88.7\%& 94.6\% & \,3.3\%\\
                     \hline
                \end{tabular}
			}
		\end{center}
  \vspace{-0.75cm}
	\end{table}

\vspace{0.1cm}
\noindent
\textbf{CIFAR-100.}
We compare our method in the more challenging setup of CIFAR-100, with 100 classes and 500 images per class, against two state-of-the-art methods: STQ \cite{munagala2020stq}, and Subbit~\cite{wang2021sub}.
We use ReActNet-18~\cite{liu2020reactnet} as the backbone architecture, using a single training step and no teacher. We train for 300 epochs with the same setup used for CIFAR-10 with Mixup augmentation. As no previous results for this setup have been reported for ReActNet-18 and Subbit, for a fair comparison, we trained them from scratch using our setup. We report the same metrics used for CIFAR-10, plus the  the reduction of binary parameters (BParams PR). For our SBNN, we estimate BParams PR as follows. For each kernel we use 2 bits to differentiate among zero Hamming weight kernels, one Hamming weight kernels and all the other kernels. Then, we add 4 bits to the kernels with Hamming weight 1 to represent the index position of their $1$-valued bit, whereas we add $9$ bits for all the other kernels with Hamming weight larger than 1, which are their original bits. For the other methods, please refer to their work for their estimate of BParams PR.

Table~\ref{tab:CIFAR_100} reports the obtained results for the different methods and our SBNN for various sparsity targets. We can see that our pruning method is more effective in reducing both the BOPs and the parameters than Subbit. It allows to remove $79.2\%$ of kernels, while increasing the original accuracy by $0.79\%$ w.r.t. the ReActNet-18 baseline. Instead, we observe nearly $1\%$ accuracy drop for a Subbit network for a similar BOPs reduction. Moreover, our method allows to remove nearly $15\%$ more binary parameters.

\begin{table*}[t]
		\caption{Evaluation of kernel removal for different pruning targets using a ReActNet-18 model on CIFAR-100.}
		\label{tab:CIFAR_100}
		\begin{center}
		  \resizebox{\linewidth}{!}{%
			\begin{tabular}{cccccrr}
			     \hline
			     \multicolumn{1}{c}{\textbf{Method}} & \multicolumn{1}{c}{\textbf{Acc.}} & \multicolumn{1}{c}{$\bm{\Delta}$} & \multicolumn{1}{c}{\textbf{BOPs PR}} & \multicolumn{1}{c}{\textbf{BParams PR}} & \multicolumn{1}{c}{$\bm{K_0}$} & \multicolumn{1}{c}{$\bm{K_1}$} \\\hline
                    ReActNet-18$^*$ & 62.79\% & / & / & / & / & /\\\cdashline{1-7}
                    STQ  & 57.72\% & -5.05\% & -36.1\% & -36.1\% & / & / \\
                    Subbit $0.67$-bits$^*$& 62.60\% & -0.19\%  & -47.2\% & -33.3\%& / & / \\
                    Subbit $0.56$-bits$^*$& 62.07\% & -0.72\%  & -70.0\% & -44.4\%& / & / \\
                    Subbit $0.44$-bits$^*$& 61.80\% & -0.99\%  & -82.3\% & -55.6\%& / & / \\
                     \rowcolor{LightCyan}
                     SBNN 50\% [\textbf{our}]  & 63.03\% & +0.24\% & -11.1\% & /& 5.6\% & 6.8\% \\
                     \rowcolor{LightCyan}
                     SBNN 95\% [\textbf{our}]  & 63.33\% & +0.54\% & -66.2\% & -59.9\%& 72.9\% & 16.6\% \\
                     \rowcolor{LightCyan}
                     SBNN 96\% [\textbf{our}]   & 63.04\% & +0.25\% & -67.3\% & -63.7\% & 78.9\% & 12.6\%\\
                     \rowcolor{LightCyan}
                     SBNN 97\% [\textbf{our}]   & 62.41\% & -0.38\% & -73.4\% & -66.8\%& 82.9\% & 11.1\% \\
                     \rowcolor{LightCyan}
                     SBNN 98\% [\textbf{our}]   & 63.58\% & +0.79\%  & -79.2\% & -70.3\%& 88.1\% & 8.0\%\\
                     \rowcolor{LightCyan}
                     SBNN 99\% [\textbf{our}]  & 62.23\% & -0.57\% & -87.8\% & -74.0\%& 93.6\% & 4.7\% \\
                     \hline
                     \multicolumn{7}{l}{ $^*$ our implementation.}
                \end{tabular}
			}
		\end{center}
  \vspace{-0.75cm}
	\end{table*}

\vspace{0.1cm}
\noindent
\textbf{ImageNet.} We assess our proposed SBNN trained with target sparsity of $75\%$ and $90\%$ on ImageNet. We compare them with state-of-the-art BNNs, namely: XNOR-Net~\cite{xnornet}, Bi-RealNet-18~\cite{liu2018birealnet} and ReActNet-18, ReActNet-A \cite{liu2020reactnet} and Subbit \cite{wang2021sub}. Moreover, we also report the accuracy of the full-precision ResNet-18~\cite{he2016resnet} and MobileNetV1~\cite{howard2017mobilenets} models, as a reference. We use a ReActNet-A~\cite{liu2020reactnet} as SBNN's backbone with its MobileNetV1 ~\cite{howard2017mobilenets} inspired topology and with the distillation procedure used in~\cite{liu2020reactnet}, whereas in Subbit \cite{wang2021sub} they used ReActNet-18 as backbone. One of the limitations of Subbit \cite{wang2021sub} is that their method cannot be applied to the pointwise convolutions of MobileNetV1~\cite{howard2017mobilenets}.
Due to  GPUs limitations, during our training, we decreased the batch size to 64. For a fair comparison, we retrained the original ReActNet-A model with our settings.

 Table~\ref{Imagenet_results_table} reports the results in terms of accuracy (Acc). We also include the number of operations (OPs) to be consistent with other BNNs assessment on ImageNet. For BNNs, OPs are estimated by the sum of floating-point operations (FLOPs) plus BOPs rescaled by a factor 1/64 \cite{xnornet,liu2018birealnet,liu2020reactnet}. We assume sparse operations on FPGA to estimate BOPs for SBNN. 
 
 We observe that BOPs are the main contributors to ReActNet-A's OPs (Table~\ref{Imagenet_results_table}), thus decreasing them largely reduces the OPs. This, instead, does not hold for ReActNet-18, which may explain why Subbit is not effective in reducing OPs of its baseline. Our method instead is effective even for less severe pruning targets and it requires less than $3.4\times$ OPs w.r.t. state-of-the-art ReActNet-A model, while incurring in an acceptable generalization loss between $1.9-3.4\%$. 

\begin{table*}[t]
		\caption{Method comparison on ImageNet.
		}
		\label{Imagenet_results_table}
		\begin{center}
		  \resizebox{.8\linewidth}{!}{%
			\begin{tabular}{l|crrr}
			\hline
				\multicolumn{1}{c|}{Model} &   \multicolumn{1}{c}{Acc} & \multicolumn{1}{c}{BOPs} & \multicolumn{1}{c}{FLOPs} & \multicolumn{1}{c}{OPs} \\
				 & \multicolumn{1}{c}{Top-1} & ($\times10^8$) & ($\times10^8$) & ($\times10^8$)\\
				\hline
				\hline
	MobileNetV1 \cite{howard2017mobilenets} (full-precision)  & $70.60$  & - & 5.7 & 5.7\\
				ResNet-18 \cite{he2016resnet} (full-precision)  & $72.12$ &   - & 19 & 19\\
				\cdashline{1-5}			
                
				{XNOR-Net \cite{xnornet}}  & $51.20$  &  17 & 1.41 & 1.67\\
				{Bi-RealNet-18 \cite{liu2018birealnet}}  & $56.40$ &   17 & 1.39 & 1.63\\
                    {ReActNet-18 \cite{liu2020reactnet}}  & $65.50$ &  17 & 1.63 & 1.89\\
				{ReActNet-A \cite{liu2020reactnet}$^*$}  & $68.12$ &  48 & 0.12 & 0.87\\
				\cdashline{1-5}
    Subbit 0.67-bits ReActNet-18  & $63.40$ &  9 & 1.63 & 1.77\\
    {Subbit 0.56-bits ReActNet-18} & $62.10$  & 5 & 1.63 & 1.71\\
    {Subbit 0.44-bits ReActNet-18}  & $60.70$ & 3 & 1.63 & 1.68\\
    \rowcolor{LightCyan}
    {SBNN 75\% ReActNet-A \textbf{[ours]}} & $66.18$ & 8 & 0.12 & 0.25\\
    \rowcolor{LightCyan}
    {SBNN 90\% ReActNet-A \textbf{[ours]}}  & $64.72$  & 2 & 0.12 & 0.16\\
				\hline
        \multicolumn{5}{l}{$^*$ our implementation.}
				
			\end{tabular}
			}
			
		\end{center}
  \vspace{-0.75cm}
	\end{table*}

\section{Conclusions}\label{sec:conclusions}

    We have presented sparse binary neural network (SBNN), a novel method for sparsifying BNNs that is robust to simple pruning techniques by using a more general binary domain. Our approach involves quantizing weights into a general $\Omega = \{\alpha, \beta\}$ binary domain that is then expressed as 0s and 1s at the implementation stage. We have formulated the SBNN method as a mixed optimization problem, which can be solved using any state-of-the-art BNN training algorithm with the addition of two parameters and a regularization term to control sparsity.

Our experiments demonstrate that SBNN outperforms other state-of-the-art pruning methods for BNNs by reducing the number of operations, while also improving the baseline BNN accuracy for severe sparsity constraints. Future research can investigate the potential of SBNN as a complementary pruning technique in combination with other pruning approaches. In summary, our proposed SBNN method provides a simple yet effective solution to improve the efficiency
of BNNs, and we anticipate that it will be a valuable addition to the field of 
binary neural network pruning.
   
%
%
\bibliographystyle{splncs04}
\bibliography{ecml.bib}
%
\section*{Ethical Statement}
\renewcommand{\labelitemi}{$\bullet$}

The proposed SBNN can in principle extend the range of devices, at the edge of communication networks, in which DNN models can be exploited. Our work touches various ethical considerations:
\begin{itemize}
\item \textbf{Data Privacy and Security}: By performing inference of DNNs directly on edge devices, data remains localized and does not need to be transmitted to centralized servers. This reduces the risk of sensitive data exposure during data transfer, enhancing privacy protection.
\item \textbf{Fairness and Bias}: SBNNs, like other DNNs at the edge, can be susceptible to biased outcomes, as they rely on training data that may reflect societal biases. However, by simplifying the weight representation to binary values, SBNNs may reduce the potential for biased decision-making because they may be less influenced by subtle variations that can introduce bias. Nevertheless, it is essential to address and mitigate biases in data to ensure fairness in outcomes and avoid discriminatory practices.
\item \textbf{Transparency and Explainability}: The SBNN design can be applied to DNN models that are designed to provide transparency and explainability. Moreover, the binary nature of SBNNs can make them more interpretable and easier to understand compared to complex, multi-valued neural networks. This interpretability can help users gain insights into the decision-making process and facilitate transparency.
\item \textbf{Human-Centric Design}: SBNNs can extend the use of DNNs at the edge, extending the range of users of applications which are focused on human well-being, human dignity and inclusivity.
\item \textbf{Resource Allocation and Efficiency}: SBNNs allows the use of DNNs in a more efficient way from both the use of energy, memory and other crucial resources, thus allowing to reduce the environmental impact of DNNs.
\item \textbf{Ethics of Compression}: While SBNNs offer computational efficiency and reduced memory requirements, the compression of complex information into binary values may raise ethical concerns. Compression may lead to oversimplification or loss of critical details, potentially impacting the fairness, accuracy, or reliability of decision-making systems.
\end{itemize}

It is important to consider these ethical aspects of SBNNs when evaluating their suitability for specific applications and to ensure responsible and ethical deployment in alignment with societal values and requirements.

\end{document}